%% file: icml2023.tex
\documentclass{article}

\usepackage{microtype}
\usepackage{graphicx}
\usepackage{booktabs} 

\usepackage{hyperref}

\usepackage[accepted]{icml2023}

\usepackage{amsmath}
\usepackage{amssymb}
\usepackage{mathtools}
\usepackage{amsthm}
\usepackage{multirow}
\usepackage[capitalize,noabbrev]{cleveref}

\theoremstyle{plain}

\theoremstyle{definition}

\theoremstyle{remark}

\usepackage[textsize=tiny]{todonotes}

\usepackage{caption}
\usepackage{subcaption}
\usepackage{bbm}
\usepackage{enumitem}

\icmltitlerunning{Few-shot Unlearning by Model Inversion}

\newcommand{\cross}{\bgroup
  \sbox0{7}\usebox0\llap{\rule[.45\ht0]{.8\wd0}{.1\ht0}}
\egroup
}
\DeclareMathOperator*{\argmax}{argmax}

\newcommand{\set}[1]{\mathcal{#1}}

\newcommand{\Oracle}{\text{Oracle}}
\newcommand{\standard}{$\Oracle{}_{\text{std}}$}
\newcommand{\correct}{$\Oracle{}_{\text{cor}}$}
\newcommand{\untrained}{$\Oracle{}_{\text{unt}}$}
\newcommand{\negative}{$\Oracle{}_{\text{neg}}$}

\newcommand{\Ours}{\text{Ours}}
\newcommand{\standardo}{$\Ours{}_{\text{std}}$}
\newcommand{\correcto}{$\Ours{}_{\text{cor}}$}
\newcommand{\untrainedo}{$\Ours{}_{\text{unt}}$}
\newcommand{\negativeo}{$\Ours{}_{\text{neg}}$}

\makeatletter
\def\blfootnote{\gdef\@thefnmark{}\@footnotetext}
\makeatother

\begin{document}

\twocolumn[
\icmltitle{Few-shot Unlearning by Model Inversion}



\icmlsetsymbol{equal}{*}

\begin{icmlauthorlist}

\icmlauthor{Youngsik Yoon}{equal,po}
\icmlauthor{Jinhwan Nam}{equal,po}
\icmlauthor{Hyojeong Yun}{po}
\icmlauthor{Jaeho Lee}{po}
\icmlauthor{Dongwoo Kim}{po}
\icmlauthor{Jungseul Ok}{po}
\end{icmlauthorlist}

\icmlaffiliation{po}{Pohang University of Science and Technology}

\icmlcorrespondingauthor{Jungseul Ok}{jungseul@postech.ac.kr}

\icmlkeywords{Machine Unlearning, Few-shot Unlearning, Model Inversion}

\vskip 0.3in
]



\printAffiliationsAndNotice{\icmlEqualContribution} 

\input{./sections/abstract}
\input{./sections/introduction}

\input{./sections/related_work}
\input{./sections/problem_formulation}

\input{./sections/proposed_method}

\input{./sections/experiments}

\input{./sections/conclusion}


\bibliography{icml2023}
\bibliographystyle{icml2023}

\newpage
\appendix
\onecolumn

\input{./sections/appendix}

\end{document}

%% file: sections/abstract.tex
\begin{abstract}

We consider a practical scenario of machine unlearning to erase a target dataset, which causes unexpected behavior from the trained model. The target dataset is often assumed to be fully identifiable in a standard unlearning scenario. Such a flawless identification, however, is almost impossible if the training dataset is inaccessible at the time of unlearning. Unlike previous approaches requiring a complete set of targets, we consider {\it few-shot unlearning} scenario when only a few samples of target data are available. To this end, we formulate the few-shot unlearning problem specifying intentions behind the unlearning request (e.g., purely unlearning, mislabel correction, privacy protection), and we devise a straightforward framework that (i) retrieves a proxy of the training data via model inversion fully exploiting information available in the context of unlearning; (ii) adjusts the proxy according to the unlearning intention; and (iii) updates the model with the adjusted proxy. We demonstrate that our method using only a subset of target data can outperform the state-of-the-art unlearning methods even with a complete indication of target data.

\end{abstract}

%% file: sections/introduction.tex
\section{Introduction}
\label{sec:intro}

Machine unlearning is the task of excising some target dataset from a trained model. The goal is to mitigate unwanted behaviors of the model that may have been induced by the target samples, which are, e.g., mislabeled \cite{cao2015towards,du2019lifelong} or privacy-sensitive (c.f., right to be forgotten \cite{mantelero2016right}). While retraining a model from scratch without using the target data is a straightforward solution, it is often impossible to do so whenever the original training data is not fully accessible due to memory or privacy concerns. To this end, \textit{standard unlearning} algorithms aim to approximate such oracle retrained model by utilizing limited \cite{golatkar2020eternal, nguyen2020variational, golatkar2020forgetting} or no \cite{fu2021bayesian} access to training data. Prior works have demonstrated that such unlearning can be successfully performed, whenever a clear and complete indication of the target data is given.

\begin{figure}[t!]
\centering
\vspace{-0.2cm}
\begin{minipage}{.45\linewidth}
\centering
\includegraphics[width=0.3\linewidth]{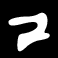}
\includegraphics[width=0.3\linewidth]{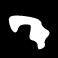}
\includegraphics[width=0.3\linewidth]{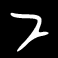}
\end{minipage}%
\hspace{0.3cm}
\begin{minipage}{.45\linewidth}
\centering
\includegraphics[width=0.3\linewidth]{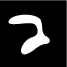}
\includegraphics[width=0.3\linewidth]{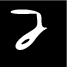}
\includegraphics[width=0.3\linewidth]{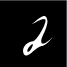}
\end{minipage}
\caption{Confusing $\cross{}$'s (left) and $2$'s (right) in MNIST.\vspace{-1em}}
\label{fig:eg}
\end{figure}

However, such a flawless indication of the target data to unlearn is not always possible. For example, consider a model that has been trained on multiple randomly augmented versions (e.g., via random cropping) of each sample. Retrieving all augmented versions of the target data used for training may not be possible unless the augmentation history has been stored during the training. As another example, we can consider online learning scenarios being fed a stream of data that is not stored due to a shortage of storage space or privacy concerns.


With the limited access to target samples, i.e., \textit{few-shot unlearning}, the desired model behavior on inaccessible samples becomes ill-specified, leading to a failure in fulfilling the intention behind the unlearning request.
For example, consider a classifier trained with a spoiled MNIST dataset of handwritten digits (0-9) where hundreds of images
of crossed-seven, denoted by $\cross{}$, are mislabeled as $2$, c.f., Figure~\ref{fig:eg}.
Given a target dataset consisting of only a {\it few} images of $\cross{}$ with label $2$,
the response to $\cross{}$ after the standard unlearning may still be $2$, as the remaining dataset after excluding only the few target data still includes hundreds of the mislabeled $\cross{}$.
Meanwhile, the unlearning requester would intend (i) a complete unlearning\footnote{Even when all the samples of $\cross{}$ are requested to be erased, the response of the retrained model is still ill-determined and possibly changing over training epochs. We refer to the Appendix~\ref{app:ill} for further discussion.}
to exclude the influence of all the mislabeled $\cross{}$; (ii) a mislabel correction to let $\cross{}$ be classified as $7$; or (iii) privacy protection to reduce the risk of recovering $\cross{}$ from the unlearned model. Standard unlearning frameworks lack an explicit mechanism to account for these intentions of the unlearner.

To address this challenge,
we first establish a generalized formulation of the few-shot unlearning
that can take into account the intention of unlearning (Section~\ref{sec:problem}). 
By specifying the intention,
our flexible formulation can correspond to 
the noisy label correction or privacy protection. Also, with a complete target indication, the formulation covers the standard unlearning as a special case.
Based on this formulation, we propose a straightforward framework (Section~\ref{section:Proposed Method})
consisting of (i) {\it model inversion}
to retrieve a proxy for training data from all the information available in the context of unlearning including the given model and target samples (Section~\ref{sec:method-inversion}); 
(ii) {\it filtration} to divide the retrieved proxy data into a part to be preserved and
the other to be unlearned or modified according to unlearner's intention while augmenting the few target samples (Section~\ref{sec:method-filtration}); and then (iii) {\it relearning} with the filtered proxy data 
to scrub out the unwanted behavior 
and impose the intended one
onto the augmented target dataset (Section~\ref{sec:method-relearning}).
For the model inversion step, we design a new inversion technique 
specialized for unlearning.
It fully utilizes the knowledge at hand in the context of unlearning: the trained model, the target samples, and the general prior on the data domain. In the relearning step, we devise a set of relearning strategies, which relabel the augmented target dataset differently from the filtration step, to impose unlearner's intention accordingly: \standardo{} for standard unlearning, 
 \untrainedo{} for privacy protection, \negativeo{} and \correcto{} for mislabeled correction.

In our experiments of
standard and few-shot unlearning setups with canonical deep learning models (Section~\ref{sec:experiment}), 
we validate the superiority of our method compared to existing ones \cite{golatkar2020eternal, chundawat2022can, Kim_2022_CVPR} with almost full access to the training and target datasets. We also demonstrate a sharp unlearning
to erase only a subclass such as \cross{}, which is infeasible in previous methods 
\cite{chundawat2022zero, baumhauer2020machine} with no access to the training dataset.
In particular, only our method can successfully unlearn a subset (e.g., \cross{}\!'s) of class.
Our empirical study suggests appropriate designs of unlearning methods for each of the canonical intentions: standard unlearning, mislabel correction, and privacy protection. This provides a unified view of existing works implicitly postulating different intentions, e.g., standard unlearning \cite{bourtoule2019machine, Kim_2022_CVPR} and privacy protection \cite{chundawat2022can}. 
In addition, the formulation with the notion of intention allows us to newly discover a practical unlearning method \negativeo{} to correct the mislabeled target dataset, applicable at no extra cost of annotations.

\smallskip
\noindent \textbf{Contributions.} The main contributions of this paper can be summarized as follows:
\begin{itemize}[leftmargin=*,topsep=0pt,parsep=-1pt]
\item We establish the generalized unlearning problem that encompasses 
not only the few-shot unlearning scenarios but also the various unlearning intentions: standard unlearning, privacy protection, and mislabel correction.

\item We devise the unlearning framework, consisting of the model inversion, filtration, and relearning steps, each of which is tailored to address few-shot unlearning with various intentions. 

\item In our experiment,
we show the strength of our method in both few-shot and standard unlearning setups with various intentions even compared to existing methods with full access to the training and target datasets. We also corroborate that 
given only a trained model and few target samples,
our method enables a sophisticated unlearning to eliminate a subclass, that is previously intractable.
        
\end{itemize}

%% file: sections/related_work.tex
\vspace{-1em}
\section{Related Work}
\label{sec:related}
\noindent \textbf{Machine unlearning.} Prior works on machine unlearning can be roughly categorized into two groups according to their assumptions on the accessibility of the training dataset (denoted by $D$) and the target dataset (denoted by $D_e \subset D$). The first group assumes full access to both $D$ and $D_e$ \cite{bourtoule2019machine, graves2020amnesiac, gupta2021adaptive, tarun2021fast, golatkar2020forgetting, chundawat2022can, Kim_2022_CVPR}, and focuses on reducing the computational cost of unlearning, to a level significantly below the cost of retraining a model from scratch using the remaining dataset $D_r := D \setminus D_e$. The second one assumes no access to $D$ except $D_e$ \cite{golatkar2020eternal, ye2022learning, nguyen2020variational,  fu2021bayesian, baumhauer2020machine, chundawat2022zero, graves2020amnesiac}. The main focus of these works is to prevent catastrophic forgetting on $D_r$ while erasing $D_e$. The works in this category typically require additional information or assumption regarding the model and the training dataset, such as
certain proxies of $D$ \cite{golatkar2020eternal, ye2022learning},
sample-wise training history \cite{graves2020amnesiac}, or Bayesian assumptions \cite{nguyen2020variational, fu2021bayesian}. Several recent works attempt to circumvent such requirements by restricting the target dataset to be \textit{all} samples of a specific class so that the remaining dataset is automatically identified by the complementary classes \cite{chundawat2022zero, baumhauer2020machine}, i.e., an exquisite unlearning to erase only a subclass (e.g., \cross{}) is not possible.
Our few-shot unlearning method needs no such additional requirement but enables the exquisite unlearning. 
\smallskip
\noindent\textbf{Model inversion.}
With growing scales of machine learning datasets and increasing attention on privacy concerns, model inversion has been widely studied in various contexts of machine learning \cite{yoo2019knowledge, chen2019data,  yin2020dreaming, choi2020data, luo2020large, yu2021conditional, zhao2022dual, fredrikson2015model, yang2019adversarial, jeon2021gradient}.
The model inversion is particularly useful in {\it data-free} transfer learning applications, including but not limited to model compression \cite{yoo2019knowledge, chen2019data, yin2020dreaming, choi2020data, luo2020large, yu2021conditional, zhao2022dual}
and continual learning \cite{yin2020dreaming},
where we want to distill the knowledge of the trained model into another but have no training data. 
In addition, to check privacy leakage, one can formulate the problem of inverting data from the model \cite{fredrikson2015model, yang2019adversarial} or even gradient \cite{jeon2021gradient}.
In this line of works on model inversion, an extensive set of prior and side information on data domain (e.g., implicit \cite{rudin1992nonlinear} and explicit \cite{ulyanov2018deep} image domain priors and batch norm statistics \cite{yin2020dreaming}) have been exploited to improve inverted data quality via reducing search space. We inherit this methodology and devise a new model inversion technique specialized under the unlearning scenario, where the target samples additionally provide hints on the training dataset. This is of independent interest to other data-free transfer learning or model inversion algorithms if few samples of training dataset are allowed.

%

%% file: sections/problem_formulation.tex
\section{Problem Formulation}
\label{sec:problem}

For ease of exposition,
we postulate a standard machine learning for image classification task
with training dataset~$D$ each of which sample is 
a pair of image $x \in \set{X}$
and label $y \in \set{Y}$, e.g., MNIST dataset of handwritten digits. Given this dataset, a model~$f(\cdot;w)$ is parameterized by~$w$
and trained by minimizing the average cross-entropy loss $\mathcal{L}(w; D)$ 
comparing $f(x_i; w)$ and $y_i$ over training dataset $D$.
We denote the parameter learned by this procedure by {\it original} model $w_o$
or equivalently $f_o := f(\cdot; w_o)$.

\smallskip
\noindent\textbf{Standard unlearning.}
Given a subset of training samples $D_e \subset D$
but limited access to training data $D$, the standard unlearning aims to tune 
the original model $w_o$ as if it has never seen $D_e$ during training. 
Meaning that, the unlearner has access to two ingredients \cite{nguyen2020variational, golatkar2020eternal}: (i) $w_o$, the original model that has been trained on the full dataset $D$; and (ii) $D_e$, the target dataset to be erased from the model.
The goal of standard unlearning is to find a model that closely approximates the behavior of the standard {\it oracle} model 
that is trained only on retained dataset $D_r:= D \setminus D_e$, i.e., 
minimizing $\mathcal{L}(w; D_r)$.

\smallskip
\noindent\textbf{Few-shot unlearning.}
In this paper, we consider a more challenging but practical scenario of few-shot unlearning
given target $D_e$ and intention $I_e$,
where $D_e$
is a subset or proxy data about the model's behavior to be erased,
and $I_e$ inheriting the input dataset of $D_e$, i.e., $(x_e, y_e) \in D_e$ if $(x_e, y'_e) \in I_e$,
describes an or no intention about the response to $D_e$ after unlearning.
The goal of few-shot unlearning is to erase an extension ${D}'_{e}$ of $D_e$ which consists of samples similar to the ones in $D_e$
and to impose the response to ${D}'_{e}$ as described in $I_e$.
In other words, 
the oracle model which our unlearning aims for
is the one
relearned from scratch using a modified training dataset ${D}'_r \cup I'_e$ where
${D}'_r := D \setminus {D}'_e$ is the remaining data after excluding samples similar to $D_e$ 
and
$I'_e$ is the extension of $I_e$ corresponding to that from $D_e$ to ${D}'_{e}$, i.e., $(x'_e, y_e) \in {D}'_{e}$ if $(x'_e, y'_e) \in I'_e$. In what follows, we describe several designs of intention $I_e$ of interests in practice, each of which defines a different notion of oracle model trained on the corresponding ${D}'_r\cup I'_e$.




\smallskip
\noindent\textbf{Intention for standard unlearning: \standard{}.}
By simply setting the empty intention, i.e., $I_e = \emptyset$, our unlearning problem becomes the standard one to erase ${D}'_{e}$.
Then, in this case, the goal is to approximate
the model trained with only ${D}'_{r}$ from scratch, called \standard{}.
The empty intention would be selected for
privacy concerns as well as correcting mislabels on the target samples.

\smallskip
\noindent\textbf{Intention for privacy protection: \untrained{}.}
Inspired by \cite{chundawat2022can}, 
we can consider an alternative intention
for privacy protection
where the intended response to ${D}'_e$
is an {\it untrained} model's output.
More formally, this intention is described as follows:
\begin{equation}
I_e = \{(x_e, y'_e):
(x_e, y_e) \in D_e,~\text{and}~
y_e' = f(x_e; w_{\text{unt}}) \} \;,\nonumber
\end{equation}
where $w_{\text{unt}}$ is a randomly initialized
parameter, and the corresponding oracle
is denoted by \untrained{}.
\untrained{} is explicitly imposed the randomized responses to ${D}'_{e}$, whereas 
\standard{} is not. 
Hence, \untrained{} would have less risk to expose ${D}'_{e}$ against inversion attacks \cite{fredrikson2015model, yang2019adversarial} to find samples of low entropy output although it might leave a watermark giving some other clues on the target samples, discussed in Section~\ref{sec:privacy}.

\vspace{-1em}
\paragraph{Intentions for mislabel correction: \correct{} and \negative{}.}
In the case that an unlearning task is requested to remove negative effects from the mislabeled dataset,
we propose two intentions at different levels of supervision.
When we can annotate the correct label for each sample in $D_{e}$ or ${D}'_{e}$,
a strong intention is the set of $(x'_e, y'_e)$'s
such that $y'_e$ is the correct label of $x'_e$, and defines \correct{}.
To save effort for the annotation of correct labels, a weak intention can be established just from knowing that $D_{e}$ is mislabeled.
As negative learning \cite{cour2011} does, it imposes the negative supervision opposite to $y_e$ upon $(x_e, y_e) \in D_e$, and defines \negative{}.

%% file: sections/proposed_method.tex
\section{Proposed Method} 
\label{section:Proposed Method}

\begin{figure*}[t!]

\centering
\includegraphics[width=0.96\linewidth]{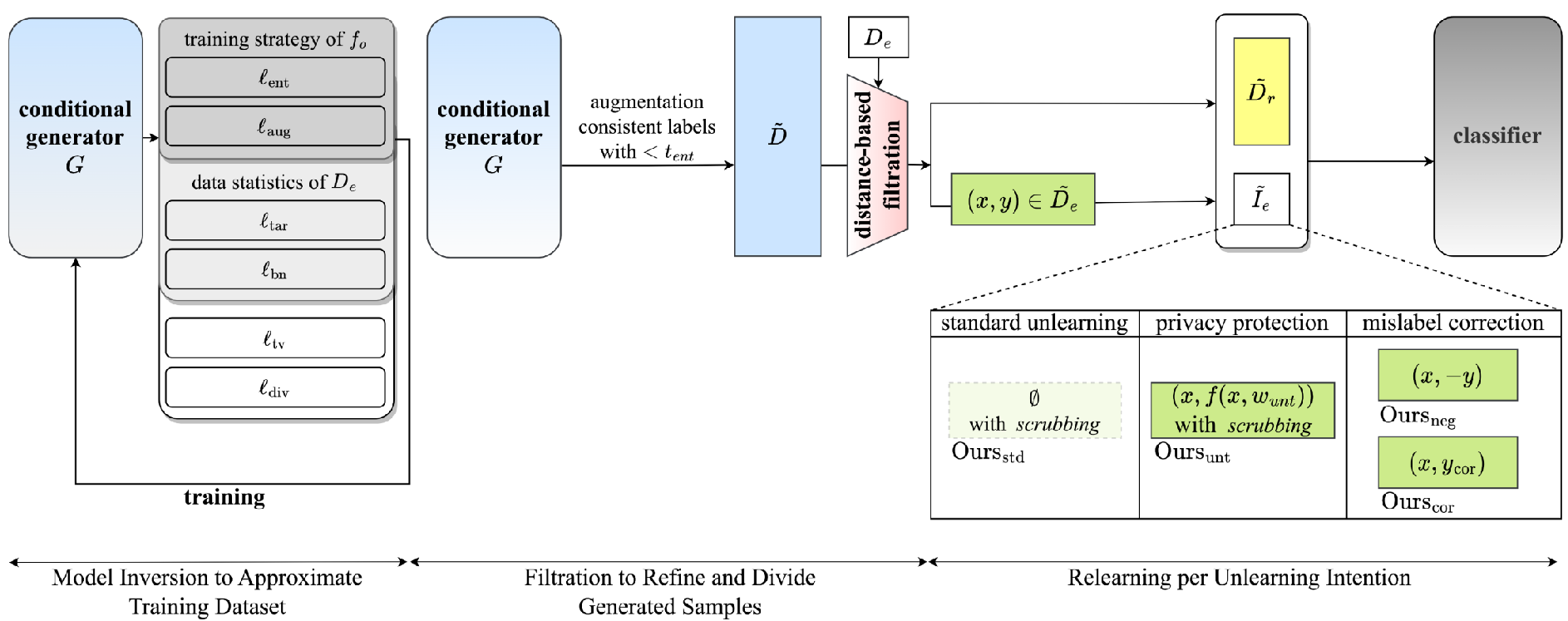}
\caption{A conceptual diagram of the proposed method.
A generator is trained by overall losses to generate $\tilde{D}$, the approximation of $D$.
The filter from $D_e$ classifies target-like samples in generated samples $\tilde{D}$, then we construct intention $\tilde{I}_e$.
The classifier unlearns specific data $D_e$ by relearning from $\tilde{D}_r \cup \tilde{I}_e $.
}
\label{figure:diagram}

\end{figure*}

The proposed algorithm consists of three steps as shown in Figure~\ref{figure:diagram}. (i) \textit{Model inversion} trains a generative model $G$ as a proxy of the original training set $D$,
where we employ an extensive set of regularizers encoding all the information at hand under the unlearning scenario, including the original model, data domain prior, and target samples. 
(ii) \textit{Filtration} first refines samples generated from $G$ to be sufficiently plausible and then partitions the refined dataset $\tilde{D}$ into 
$\tilde{D}_r$ and $\tilde{D}_e$, which are the sets of $D_r$-like and $D_e$-like samples, respectively. In this process, the target dataset $D_e$ is augmented.
(iii) \textit{Relearning} prepares $\tilde{I}_e$ by relabeling $D_e$-like samples correspondingly to an unlearning intention and then adjusts the model $f$ after scratching the original model $f_o$'s response to target samples out.
In the remainder of this section, we describe each step of the algorithm in more detail. 


\subsection{Model Inversion to Approximate Training Dataset}
\label{sec:method-inversion}

Recalling we have only access to the original model $f_o$ and 
few targets $D_e$, we need to retrieve a proxy of the entire dataset $D$ from $f_o$ but also to interpolate $D_e$ for the augmented target dataset $D'_e$.
To this end, let $\set{Y}^+$ be an extension of $\set{Y}$ with an additional auxiliary class $y^+$\footnote{For simplicity, we assume $D_e$ consisting of a single label~$y^+$.}
where the $D_e$-like samples are designated by class $y^+$.
Then, we employ a conditional generative model $G:\mathcal{Z}\times \mathcal{Y}^{+} \mapsto \mathcal{X}$ such that, given 
a random noise $z \in \set{Z}$ and class $y \in \mathcal{Y}^{+}$,
$G(y, z)$ is a randomly generated image $x \in \set{X}$ 
that is similar to (i) the target samples in $D_e$ if $y=y^+$;
or (ii) the samples participated in the training of $w_o$ with label $y$ otherwise.
To train such a conditional generator $G$,
we 
utilize an extensive set of loss functions ($\ell_{\text{tv}}, \ell_{\text{div}}, \ell_{\text{ent}}, \ell_{\text{aug}}, \ell_{\text{tar}}$, and $\ell_{\text{bn}}$)
to be minimized,
where $\ell_{\text{tar}}$ is used only for the auxiliary class $y^+$.
$\ell_{\text{tv}}$ captures our general belief on natural images.
 $\ell_{\text{div}}$ is a regularizer, borrowed from \cite{yoo2019knowledge}, to prevent
the mode-collapse issues and to diversify the generator's output images.
The formal definitions of $\ell_{\text{tv}}$ and $\ell_{\text{div}}$ are given in Appendix~\ref{app:def}, where we also leave a suggestion to further improve the quality of inverted samples based on the deep image prior \cite{ulyanov2018deep}. In what follows, we describe the remaining losses that are induced from the training strategy of $f_o$ and the data statistics of $D_e$ or $D$.

\smallskip
\noindent\textbf{$\ell_{\text{ent}}$ and $\ell_{\text{aug}}$ from the training strategy.}
Presuming a fairly successful training of $f_o$, 
a training sample $(x,y) \in D$ is anticipated to 
have a low cross entropy between the prediction $f_o(x)$ and label $y$ in most cases.
We hence employ 
$\ell_{\text{ent}}$ 
comparing the predicted label $f_o(\tilde{x})$ and the conditional label $y$ of generated sample $\tilde{x} = G(z, y)$
in terms of cross entropy.
It encourages the generated sample $\tilde{x} = G(z, y)$
to be predicted as the label of $D_e$ if $y = y^+$
or the class $y$ otherwise. 
From the fair training of $w_o$,
we can also assume that the classifier $f_o$ is robust against a set $\Phi$ of data augmentation $\phi: \set{X} \mapsto  \set{X}$ such as flipping, rotating, and cropping, used in a training step.
Indeed, such augmentations are often included in the training strategy of $f_o$. 
In other words, 
if the generated samples are similar to the training ones, then the classification result of $f_{o}$ on the original and augmented samples need to be similar to each other.
Otherwise, they are unlikely used in training $f_{o}$.
From this, we define a data augmentation loss $\ell_{\text{aug}}$ 
for 
each generated samples $\tilde{x} = G(z, y)$ as follows:
\begin{equation}
    \ell_{\text{aug}} (\tilde{x}) = 
    \sum_{\phi \in \Phi}
    \lVert{f_o(\tilde{x}) - f_{o}(\phi (\tilde{x}))}\rVert_{2}^{2} \;,
\end{equation}
where the augmentation $\phi$ is possibly random.

\smallskip
\noindent\textbf{$\ell_{\text{tar}}$ and $\ell_{\text{bn}}$ from the data statistics.}
To generate target-like samples, we can directly 
exploit the statistics about the given target $D_e$.
$\ell_{\text{tar}}$ is a discrepancy measure between
a set $B^+$ of generated {\it target-like} sample $G(z, y^+)$ and
$D_e$. Specifically, the discrepancy is defined in the latent feature space of $f_o$
rather than the image space as we want to generate images semantically similar to $D_e$.
More formally, let $f_{o, l}$ be
the intermediate process up to the $l$-th layer of $f_o$. Then,
for the mini-batch $B^+$ of generated {\it target-like} sample $\tilde{x}^+ = G(z, y^+)$, 
we define 
\begin{align} \label{eq:loss-tar}
    \ell_{\text{tar}}(B^+) = 
    \sum_{l} 
\lVert{ \mu_{l} (B^+) - \mu_{l} (D_e)}\rVert_{2}^{2} 
         \;,
\end{align}
where
the summation is taken over the layers of interests
and
$\mu_{l} (B) := \frac{1}{|B|} \sum_{x \in B} f_{o,l}(x)$
is the average of the $l$-th intermediate outputs over some image set $B$.
We can let $\ell_{\text{tar}}$ compare the variances of intermediate outputs 
but we do not as the empirical variance from the few samples is prone to distorted.

Besides the statistics of the target dataset $D_e$,
that of the entire dataset $D$ is often available.
Indeed, 
the use of the batch normalization (BN) layers is a de-facto standard architecture of deep classifier,
and each BN layer represents a meta-data of $D$ in terms of the running mean $\mu_{l}(D)$ and variance $\sigma^2_{l}(D)$.
Hence, when the given model $f_o$ includes such BN statistics,
we can utilize them as proposed in \cite{yin2020dreaming}.
Specifically, the following BN loss $\ell_{\text{bn}}$ forces generated images to fit the statistics of $D$: for a set $B$ of generated images $\tilde{x} = G(z, y)$ 
with $y$ drawn from $\set{Y}^+$, 
\begin{align*}
    \ell_{\text{bn}}(B) =  \!\!\!
    \sum_{l: \text{BN layers}} \!\!\!
                \lVert \mu_{l}(B) - \mu_{l}(D)\rVert_{2}^{2} + 
                \lVert \sigma_{l}^{2}(B) - \sigma_{l}^{2}(D) \rVert_{2}^{2}
         \;,
\end{align*}
where $\sigma^2_l(B)$ is the running variance of the $l$-th layer over $B$.
We remark that
$\ell_{\text{bn}}$ 
guides the generation of samples over all the conditions $\set{Y}^+$ (even including $D_e$),
whereas it is available only when the classifier includes BN layers.
However, $\ell_{\text{tar}}$ is applicable regardless of the existence of BN layers
and thus the summation of $\ell_{\text{tar}}$ in \eqref{eq:loss-tar} can take over any layer set, while that of $\ell_{\text{bn}}$ is restricted to BN layers.
Also,
although the knowledge of $\ell_{\text{tar}}$ seems just specialized to only the generation of target-like samples, our experiment suggests that the conditional generator can transfer the knowledge across all the classes $\set{Y}$ (Table~\ref{table:ablation-loss}).

\subsection{Filtration to Refine and Divide Generated Samples}
\label{sec:method-filtration}

\smallskip
\paragraph{Filtration for $\tilde{D}$.}
We construct the approximated dataset $\tilde{D}$ from samples of the trained generator $G$.
As we enforce the diversity of outputs from $G$,
some of the samples might not be close to any image in the original dataset $D.$
To further refine the generated images, we only incorporate samples whose (i) classification entropy with augmentation are lower than a pre-defined threshold $t_{\text{ent}}$; and (ii) predicted labels are consistent with and without augmentation.
For those samples satisfying the two criteria, we use the predicted probability as a soft label for the generated image.
Formally, the approximated dataset $\tilde{D}$ consists of $(\tilde{x}, f_{{o}}(\tilde{x}) )$
such that: for each augmentation $\phi \in \Phi$,
\begin{align*}
&
H(f_{{o}}(\Tilde{x})) < t_{\text{ent}}, \quad
H(f_{{o}}(\phi(\Tilde{x}))) < t_{\text{ent}}, \quad \text{and} \\
&\argmax_{y} f_{{o}}(\phi(\Tilde{x})) = \argmax_{y} f_{{o}}(\Tilde{x}) \;,
\end{align*}
where the entropy $H(f_{{o}}(\Tilde{x}))$ is a measure of certainty about the prediction $f_o(\Tilde{x})$.




\smallskip
\noindent\textbf{Filtration for $\tilde{D}_e$ and $\tilde{D}_r$.}
Although the sample generation of $G$ can be conditioned 
on $y^+$ for target-like samples,
we cannot partition $\tilde{D}$ into $\tilde{D}_r$ and $\tilde{D}_e$
just relying on the labels used in the generation.
We hence devise a distance-based filtration to identify target-like samples.
Specifically, we measure the distance between the generated images and the images in $D_{e}$ in the latent space at the penultimate layer $f_{o,*}$ of $f_o$ instead of the input space, and then remove the images under a certain distance threshold $t_\text{f}$.
Formally, the approximation $\tilde{D}_e$ of $D_e$ consists of $(x, y) \in \tilde{D}$
such that:
\begin{align} \label{eq:filter}
\quad \kappa(x, D_{e}; f_{o,*}) < t_{\text{f}} \;,
\end{align}
where 
it directly defines $\Tilde{D}_{r} = \Tilde{D} \setminus \Tilde{D}_{e}$ 
and
$\kappa$ is a distance metric,
measuring distance between a given sample $x$ and $D_{e}$ by maximum mean discrepancy (MMD) score. RBF kernel is employed for MMD score in this work.
Noting that the nature of the exponential in RBF kernel of $\kappa$
provides a sudden change of $\kappa$ around the neighborhood of any point in $D_e$
on the feature map,
we have a clear distinction of $\kappa$ values for target-like samples and others,
c.f., Figure~\ref{figure:model_inversion_sorted}.
Based on this observation, we use a knee point detection mechanism \cite{satopaa2011finding}
to automatically determine a proper threshold $t_{\text{f}}$ instead of handcrafting it.




\begin{table*}[!ht]
\centering
\begin{tabular}{lcccccc}
\toprule
\multirow{2}{*}{Method} 
& \multicolumn{2}{c}{{\it \cross\!-noisyMNIST} (\%)}
& \multicolumn{2}{c}{{\it \cross\!-MNIST} (\%)} 
& \multicolumn{2}{c}{{\it Truck-CIFAR10} (\%)}\\
\cmidrule(lr){2-7}
 & \multicolumn{1}{c}{$D_r (\uparrow)$}  & \multicolumn{1}{c}{$D_e (\uparrow)$}
 & \multicolumn{1}{c}{$D_r (\uparrow)$}  & \multicolumn{1}{c}{$D_e$}
 & \multicolumn{1}{c}{$D_r (\uparrow)$}  & \multicolumn{1}{c}{$D_e$} 
 \\
\hline
 Original     
 & $99.4$ & $ 0.0$
 & $99.7$ & $98.5$ 
 & $92.7$ & $96.3$
 \\
 \cmidrule(lr){1-7}
 \standard{}  
 & $99.5$ & $74.5$
 & $99.5$ & $74.5$ 
 & $92.9$ & $ 0.0$\\
 \untrained{} 
 & $99.5$ & $19.0$
 & $99.6$ & $19.0$
 & $93.3$ & $ 0.0$
 \\
 \negative{}  
 & $99.6$ & $100.0$
 & $98.1$ & $ 0.0$ 
 & $93.0$ & $0.0$\\
 \correct{}   
 & $99.5$ & $100.0$
 & \_      & \_      
 & \_       &\_\\ 
 \cmidrule(lr){1-7}
 Incompetent 
 & $99.5_{\pm 0.5}$ & $91.0_{\pm 1.9}$
 & $99.5_{\pm 0.1}$ & $17.4_{\pm 7.8}$ 
 & $90.3_{\pm0.7}$  & $0.7_{\pm1.6}$\\
 Neutralized 
 & $99.5 _{\pm 0.1}$ & $48.6 _{\pm 24.9}$
 & $99.7 _{\pm 0.0}$ & $99.5 _{\pm 0.0}$ 
 & $92.9_{\pm0.2}$ & $1.5_{\pm0.4}$\\
 Fisher 
 & $97.4 _{\pm 0.9}$ & $ 0.9 _{\pm 0.6}$
 & $52.9 _{\pm 12.4}$ & $ 8.4 _{\pm 13.6}$ 
 & $87.5_{\pm1.4}$ & $87.4_{\pm6.2}$ \\
 \midrule
 \standardo{}    
 & $98.9 _{\pm 0.2}$ & $ 7.0 _{\pm 3.5}$
 & $99.3 _{\pm 0.3}$ & $81.5 _{\pm 20.1}$ 
 & $86.4_{\pm0.5}$ & $1.6_{\pm1.0}$ \\
 \untrainedo{}   
 & $98.4 _{\pm 0.3}$ & $33.3 _{\pm 6.0}$
 & $98.3 _{\pm 0.5}$ & $31.1 _{\pm 10.3}$ 
 & $87.1_{\pm0.7}$ & $0.4_{\pm0.2}$ \\
 \negativeo{}    
 & $99.1 _{\pm 0.0}$ & $100.0_{\pm 0.0}$
 & $98.5 _{\pm 0.5}$ & $ 2.0 _{\pm 0.6}$ 
 & $89.4_{\pm0.3}$ & $2.3_{\pm0.6}$ \\
 \correcto{}    
 & $99.3 _{\pm 0.0}$ & $100.0_{\pm 0.0}$   
 & \_                  & \_                  
 & \_                  & \_\\
 \cmidrule(lr){1-7}
 \standardo{} (3\%)   
 & $98.5 _{\pm 0.9}$ & $ 5.6 _{\pm 2.7}$
 & $99.6 _{\pm 0.1}$ & $90.8 _{\pm 2.6}$ 
 & $86.4_{\pm0.7}$ & $4.5_{\pm1.3}$ \\
 \untrainedo{} (3\%)   
 & $97.6 _{\pm 0.9}$ & $32.4 _{\pm 15.5}$
 & $98.7 _{\pm 0.1}$ & $52.9 _{\pm 14.5}$ 
 & $86.9_{\pm0.2}$ & $0.7_{\pm0.6}$ \\
 \negativeo{} (3\%)   
 & $99.3 _{\pm 0.1}$ & $99.7 _{\pm 0.4}$
 & $98.1 _{\pm 0.1}$ & $11.7 _{\pm 0.5}$ 
 & $89.7_{\pm0.8}$ & $2.3_{\pm1.4}$ \\
 \correcto{} (3\%)    
 & $99.3 _{\pm 0.0}$ & $100.0_{\pm 0.0}$
 & \_                  & \_                  
 & \_                  & \_\\
\bottomrule
\end{tabular}
\caption{Test accuracies on $D_{r}$ and $D_{e}$ of unlearning methods
on the representative scenarios. 
We report the mean with standard deviation over 5 random instances.
The results in {\it \cross{}\!\!-noisyMNIST}, {\it \cross{}\!\!-MNIST}, and
{\it Truck-CIFAR10} are mainly discussed in Section~\ref{sec:mislabel}, \ref{sec:privacy}, and \ref{sec:justify}, respectively.
}
\label{table:mnist}
\end{table*}

\subsection{Relearning per Unlearning Intention}
\label{sec:method-relearning}

We so far obtained $\tilde{D}_e$ and $\Tilde{D}_{r}$.
Noting that the few-shot unlearning is defined with a specific intention
as described in Section~\ref{sec:problem}, 
we need to further prepare $\tilde{I}_e$ by relabeling $\tilde{D}_e$ with the label 
$y'$ corresponding to the intention, i.e.,
\begin{equation}
\tilde{I}_e := \big\{(x, y'): (x, y) \in \Tilde{D}_{e}\big\} \;.
\end{equation}
Now, we describe how to implement our unlearning method
for each intention associated with an oracle in Section~\ref{sec:problem}:
\begin{itemize}[leftmargin=*,topsep=0pt,parsep=-1pt]
\item \standardo{}. For the standard unlearning, we can blindly imitate \standard{}
and fine-tune from $f_o$ with only $\tilde{D}_r$ discarding $\tilde{D}_e$.
However, the response to target samples can be significantly biased by $f_o$ and
different from \standard{}. 
Analogously to \cite{Kim_2022_CVPR},
we have a {\it scrubbing procedure} in a few early epochs
that learns $\Tilde{D}_{e}$ with random labels to quickly erase $f_o$'s response to $\tilde{D}_e$ before fine-tuning with only $\Tilde{D}_{r}$. 
The importance of the scrubbing is investigated 
via an ablation study in Appendix~\ref{app:scrub}.
The aforementioned procedure forms \standardo{}.

\item \untrainedo{}. For privacy protection, 
\untrained{} intended to remove target data by pseudo-labeling their label 
from the randomly initialized model.
In this case, we also use the output of a randomly initialized model $w_\text{unt}$ as a pseudo-label of $\tilde{D}_e$ for \untrainedo{} after a few epochs of the scrubbing procedure at the beginning to accelerate the unlearning.
We note that it is possible to consider an unlearning with only the scrubbing procedure for privacy protection. However, this is unfortunately vulnerable to the membership inference attack \cite{shokri2017membership} because it often
imposes a low entropy of the response to a target sample, which is conspicuous to attackers.

\item \negativeo{}. In the case of pursuing \negative{},
we use the negative learning ({\it minus} target label), which tries to make zero prediction probability at the label of target. In this case, we have an indirect yet specific intention on $\tilde{D}_e$. Hence, \negativeo{} skips the scrubbing
and just fine-tune the model with $\tilde{D}_r$ and $\tilde{I}_e$ with the negative labels.

\item \correcto{}. For \correct{} with the knowledge of correct
labels of $D_e$, 
we assign the correct label $y'$ for each target sample in $\tilde{I}_e$.
Similarly with \negativeo{},
\correcto{} directly performs couples of fine-tuning steps with $\tilde{D}_r$ and $\tilde{I}_e$ with the corrected labels, while the label correction can be expensive in practice.


\end{itemize}



%% file: sections/experiments.tex
\newcommand{\cmark}{\textcolor{black!80!black}{\ding{51}}}

\section{Experiment}
\label{sec:experiment}




% ...

\noindent
\textbf{Scenarios.}
In this section, we present the numerical analysis in the following three main unlearning scenarios, 
where canonical deep neural networks are trained for image classification based on MNIST \cite{lecun1998gradient} and CIFAR-10 \cite{krizhevsky2009learning} datasets.
Unlearning tasks are requested with 100\% (full-shot) or 3\% (few-shot) of $D_e$, where we omit 100\% in cases of full-shot unlearning for notational simplicity.
Detailed setups and more experiments carrying similar messages are postponed to the appendix.
\begin{itemize}[leftmargin=*,topsep=0pt,parsep=-1pt]

\item {\it \cross{}\!-noisyMNIST} scenario
uses noisyMNIST where we modify MNIST
such that all the samples in subclass \cross{} are mislabeled as class $2$.
A model is trained on noisyMNIST, and the target dataset $D_e$ is the mislabeled subclass {\cross{}}.

\item {\it \cross{}\!-MNIST} scenario starts with 
a model trained on MNIST. 
The unlearning task is requested to erase a subclass \cross{} in class $7$, i.e.,
$D_e$ is the subclass \cross{}.

\item {\it Truck-CIFAR10} scenario begins with 
a model trained on CIFAR10.  
We want to erase the whole of {\it truck} class, while maintaining the other classes. This is one of the simplest unlearning scenarios. However, it is much more challenging than a similar one based on MNIST (grayscale) due to its higher complexity of data domain (colored) in particular when we have no free access to $D$.


\end{itemize}

\smallskip
\noindent
\textbf{Baselines.}
We compare the models from a set of unlearning methods, described in what follows:
\begin{itemize}[leftmargin=*,topsep=0pt,parsep=-1pt]
\item {\it Original} denotes the model $f_o$ trained with $D$ in advance of unlearning, i.e., no unlearning.

\item {\it Oracle} is an ideal reference model
trained with complete $D_r$ and modified $D_e$ for each task from the scratch,
with some specific intention as described in Section~\ref{sec:problem}.

\item {\it Ours}
is from our few-shot unlearning method
with the generator trained from scratch
in Section~\ref{sec:method-inversion}.
An unlearning method can be numerically assessed 
by measuring the discrepancy between the behaviors of the oracle and its output model,
whereas a better unlearning method has a smaller discrepancy to the oracle of the context of interests, e.g., for mislabel correction, \correct{} would be compared to 
\correcto{} or \negativeo{}.





\item {\it Incompetent}, {\it Neutralized}, and {\it Fisher},
are the unlearning methods proposed by  \citet{chundawat2022can}, \citet{Kim_2022_CVPR}, and \citet{golatkar2020eternal}, respectively, and 
dedicated to \untrained{}, \standard{}, and \standard{}, respectively.
We note that they require practically free access to $D$ although they 
are applicable to canonical deep neural networks.

\end{itemize}

\subsection{Mislabel Correction}
\label{sec:mislabel}

\begin{figure*}[t!]
    \begin{subfigure}{0.57\linewidth}
    \centering
        \includegraphics[width=0.42\linewidth]{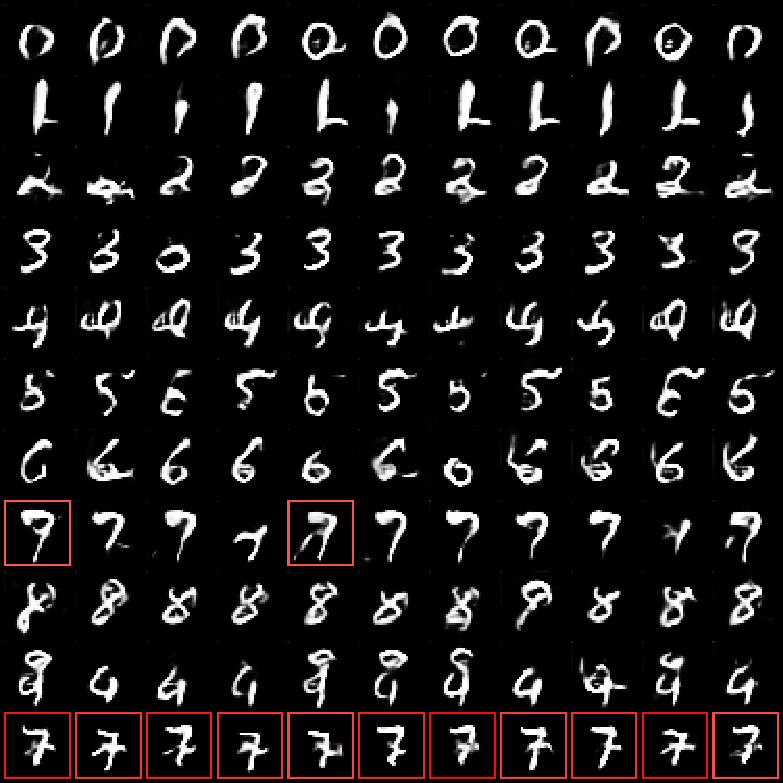}
        \includegraphics[width=0.536\linewidth]{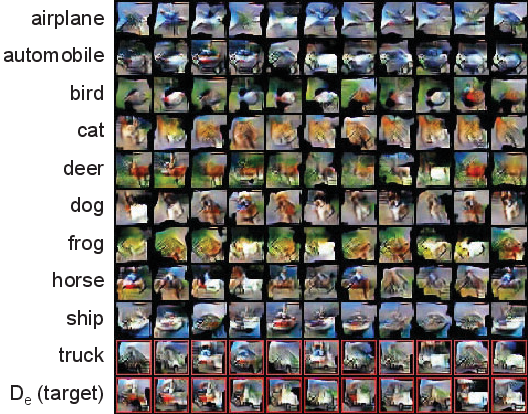}
        \caption{}
        \label{figure:model_inversion}
    \end{subfigure}
    \begin{subfigure}{0.41\linewidth}
    \centering
        \includegraphics[width=0.47\linewidth]{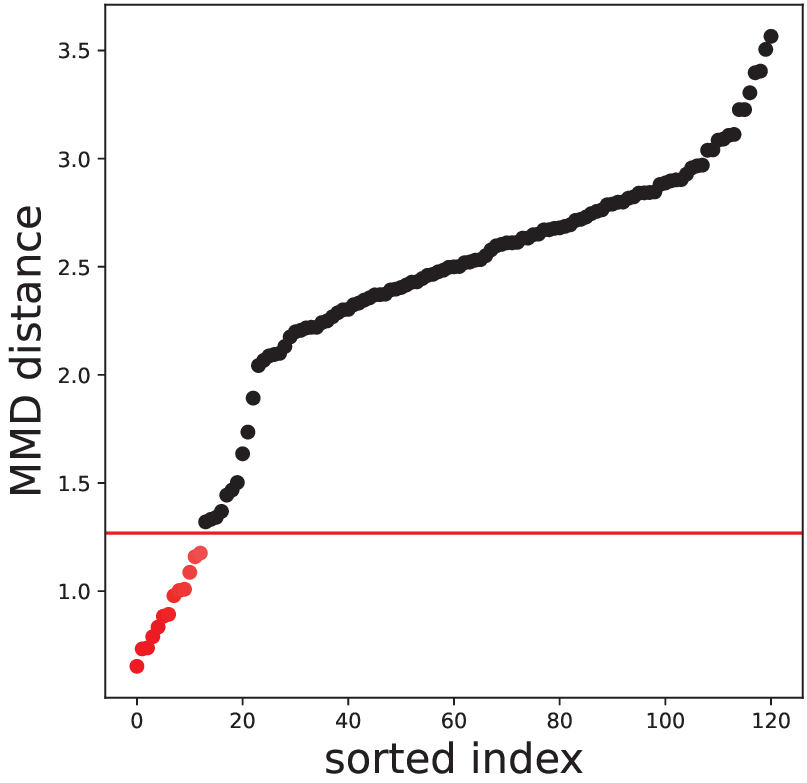}
        \includegraphics[width=0.47\linewidth]{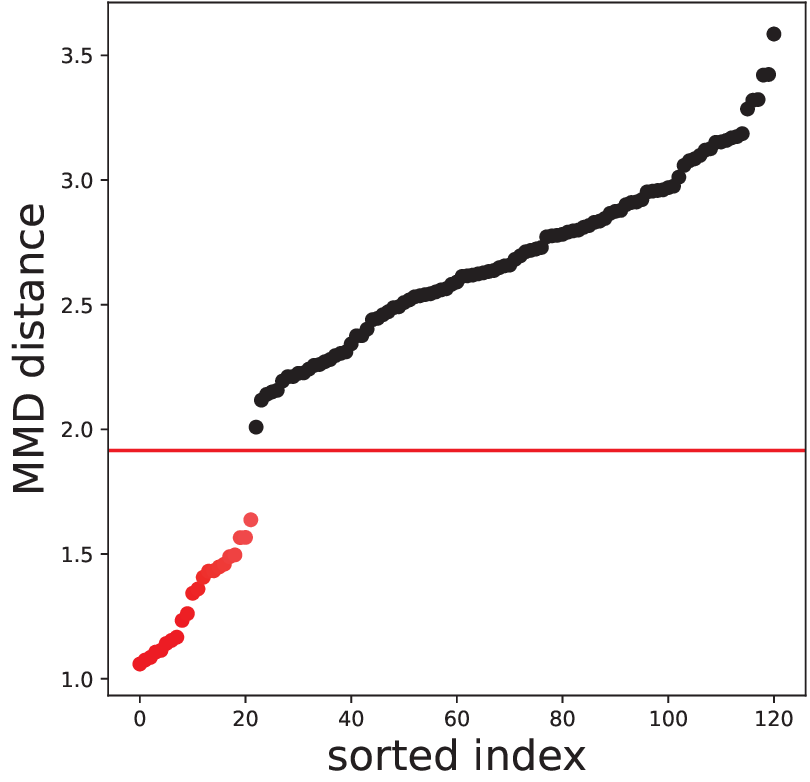}
        \caption{}
        \label{figure:model_inversion_sorted}
    \end{subfigure}
\caption{
(a) Generated images in $\tilde{D}$, where the target-like samples $\tilde{D}_e$ identified by the filtration step 
are marked with the red boxes;
and (b) sorted MMD of a generated sample in $\tilde{D}$ to $D_e$,
where 
the horizontal line indicates the threshold $t_{\text{f}}$ for the filtration found by knee point detector. 
(left: \textit{\cross{}\!-MNIST}; and right: \textit{Truck-CIFAR10})
}
\label{figure:MNIST_D_r_D_e}
\end{figure*}

\noindent
\textbf{\correcto{} and \negativeo{} for correction.}
The first column of Table~\ref{table:mnist} compares unlearning methods under {\it \cross{}\!-noisyMNIST} scenario
of which the ideal objective
is \correct{} that corrects Original's misbehavior induced from the mislabeled subclass~\cross{}.
The objective of \correct{} is closely achieved by both \correcto{} and \negativeo{},
which drastically improved the low accuracy of Original on $D_e$ 
while maintaining the high accuracy on $D_r$.
Remarkably, \correcto{} and \negativeo{} given only 3\% of $D_e$ (24 images of \cross{}\!)
also attain the ideal model correction, closely.
Recalling \correcto{} requires additional efforts to collect the correct annotations of $D_e$ but \negativeo{} does not,
this suggests that \negativeo{} is a particularly practical unlearning method for mislabel correction.

\noindent
\textbf{Necessity of more than erasing for correction.}
In the experiment of {\it \cross{}\!-noisyMNIST},
we also observe a substantial gap between \correct{} and \standard{} (or others dedicated to it: Incompetent, Fisher, and \standardo{}) 
even if all the mislabeled target samples are indicated.
This gap implies that more than just erasing is necessary to remedy the misbehavior induced by $D_e$. Otherwise, the response on $D_e$ becomes unpredictable after unlearning, c.f., the oscillating response to $D_e$ of \standard{} in Appendix~\ref{app:ill}.


\subsection{Privacy Protection}
\label{sec:privacy}

The second column of Table~\ref{table:mnist} provides a comparison of
unlearning methods in {\it \cross{}\!-MNIST}
of which the unlearning request can be asked for privacy protection of subclass \cross{}\!. An ideal unlearning result is to retain Original's accuracy on $D_r$ (including non-crossed 7's) and to have an equivocal prediction on $D_e$ (subclass \cross{}) so that the samples in $D_e$ are not easily identifiable by the model response. In this sense, only Neutralized fails to hide $D_e$ since its accuracy on $D_e$ is as high as Original. However, this analysis is insufficient to identify suitable unlearning methods for preserving privacy.



\noindent
\textbf{Membership inference attack.}
To further investigate unlearning to protect data privacy, we borrow a measure of privacy leakage
which is {\it attack success rate} (ASR) of a membership inference method \cite{shokri2017membership}, also used in \cite{golatkar2020forgetting}.
To be specific,
a binary support vector (SVM) taking the model $f$'s output to a sample
is trained to infer whether or not the sample is used in the training of $f$. 
Noting that a privacy leakage can be conducted by model inversion 
to retrieve inputs having a {\it distinguishable} output of $f$ (one-hot  or low-entropy vector),
we can anticipate that the SVM identifies the samples that have $f$'s distinguishable outputs implying high risks of model inversion attack success. 
In this sense, 
low ASR on $D_e$ is a necessary condition for preserving privacy on $D_e$.


\noindent
\textbf{Necessity of more than erasing (\untrainedo{}) for privacy.}
Table~\ref{table:privacy} first compares the ASR's of unlearning methods in {\it \cross{}\!-MNIST} scenarios. This comparison suggests \untrained{} as a reference oracle for privacy protection. Hence, 
Incompetent and \untrainedo{} show low ASR's on $D_e$
as they aim at imitating \untrained{}. 
Again, \untrainedo{} fairly protects $D_e$'s privacy even when 
3\% of $D_e$ is given. 
As expected from 
the analysis based on accuracy, Neutralized has a high ASR.
It is also interesting that \standard{} completely fails while \standardo{} 
reduces ASR on $D_e$. However, the huge gap between ASRs of 
\untrainedo{} and \standardo{} verifies the necessity of more than just erasing
for privacy protection.

\noindent
\textbf{Watermark after unlearning.}
In Table~\ref{table:privacy}, 
we also report L2 norm average of the model $f$'s penultimate outputs to 
$D_r$ and $D_e$. Interestingly,
\untrained{}, \untrainedo{} and Incompetent,
which successfully make the (ultimate) outputs to $D_e$ indistinguishable,
have noticeably small L2 norm on $D_e$, i.e., the penultimate outputs to $D_e$ around the origin. 
From this fact, it is possible to design a new model inversion attack
to retrieve the unlearned samples by finding input data such that its feature on
the penultimate layer is close to the origin, i.e.,
the untrained intention carves a {\it watermark} in the penultimate layer.

\begin{table}[t!]
\centering
\begin{tabular}{lcccc}
\toprule
\multirow{2}{*}{Method} & \multicolumn{2}{c}{ASR (\%)} & \multicolumn{2}{c}{L2 norm}\\
\cmidrule(lr){2-5}
 & \multicolumn{1}{c}{$D_r$}  & \multicolumn{1}{c}{$D_e (\downarrow)$} 
 & \multicolumn{1}{c}{$D_r$}  & \multicolumn{1}{c}{$D_e$} 
 \\
\hline
 Original                                 
 & $100.0$              & $100.0$       
 & $20.6_{\pm3.4}$      & $21.9_{\pm3.2}$       \\
 \cmidrule(lr){1-5}
 \standard{}       
 & $100.0$              & $100.0$       
 & $13.5_{\pm1.0}$      & $12.5_{\pm0.8}$       \\
 \untrained{}       
 & $99.5$               & $0.0$       
 & $12.3_{\pm0.9}$      & $7.7_{\pm0.2}$       \\
 \negative{}       
 & $100.0$              & $100.0$       
 & $13.0_{\pm0.9}$      & $12.0_{\pm0.4}$       \\
 \cmidrule(lr){1-5}
 Incompetent 
 & $99.4$               & $23.8$       
 & $24.0_{\pm14.6}$     & $6.1_{\pm0.9}$       \\
 Neutralized 
 & $100.0$              & $100.0$       
 & $23.2_{\pm8.9}$      & $20.0_{\pm3.0}$       \\
 Fisher
 & $100.0$              & $100.0$
 & $19.5_{\pm3.2}$      & $17.7_{\pm2.4}$   \\
 \midrule
 \standardo{}         
 & $99.7$               & $82.5$       
 & $19.2_{\pm4.9}$      & $11.2_{\pm1.2}$       \\
 \untrainedo{}         
 & $99.7$               & $30.0$       
 & $17.4_{\pm6.3}$      & $4.2_{\pm1.0}$       \\
 \negativeo{}         
 & $99.8$               & $100.0$       
 & $21.7_{\pm4.3}$      & $19.3_{\pm2.8}$       \\
 \cmidrule(lr){1-5}
 \standardo{} (3\%)  
 & $100.0$              & $85.0$       
 & $20.0_{\pm4.4}$      & $12.4_{\pm1.8}$       \\
 \untrainedo{} (3\%)  
 & $100.0$              & $32.5$       
 & $18.0_{\pm6.6}$      & $4.1_{\pm1.2}$       \\
 \negativeo{} (3\%)  
 & $100.0$              & $100.0$       
 & $21.8_{\pm4.2}$      & $18.8_{\pm2.5}$       \\
\bottomrule
\end{tabular}
\caption{Attack success rate (ASR) and averaged L2 norm of the penultimate features in \textit{\cross\!-MNIST}.
Attack Success Rate reports the result of the membership inference attack method.
L2 norm shows watermark effect mentioned in Section~\ref{sec:privacy}, that depending on unlearning method, $D_e$'s norm in latent space can be significantly small which is an evidence that such samples are unlearned.
}
\label{table:privacy}
\end{table}

\subsection{Design Justification}
\label{sec:justify}
In what follows, we provide empirical justifications for our design choices.

\noindent
\textbf{Challenging model inversion.}
The model inversion of {\it Truck-CIFAR10} is more challenging
than that of {\it \cross{}\!-MNIST} due to the increased complexity of input domain (grayscale $\to$ RGB). 
However, as shown in Figure~\ref{figure:MNIST_D_r_D_e}, our inversion method produces 
reasonable proxies for both {\it \cross{}\!-MNIST} and {\it Truck-CIFAR10}.
Also, the last column of Table~\ref{table:mnist} 
shows that even in {\it Truck-CIFAR10} and given a few targets, 
our unlearning method with each intention successfully unlearns
$D_e$ as much as the corresponding oracle does, although
the accuracy on $D_r$ is slightly degenerated after unlearning.
The quality and diversity of images generated by the inversion could be one reason for the performance degeneration.

\noindent
\textbf{Key loss in model inversion.}
Recalling that in {\it \cross{}\!-noisyMNIST} we can assess
the quality of unlearning clearly in terms of the accuracy on $D_r$
and $D_e$ (higher is better), 
Table~\ref{table:ablation-loss} presents
an ablation study of the key losses 
($\ell_\text{tar}$, $\ell_\text{aug}$, and $\ell_\text{bn}$)
in our model inversion method
by evaluating the downstream unlearning performance.
Apparently, $\ell_\text{tar}$, which is a novel loss specialized for the unlearning context, covers the most gain from 
$\ell_\text{aug}$ and $\ell_\text{bn}$.
In a typical model inversion, $\ell_\text{bn}$ is known to be the most prominent \cite{yin2020dreaming}.
Hence, the ablation study suggests that even when 
$\ell_\text{bn}$ is inapplicable due to no BN layers,
we can get similar results by actively utilizing
$\ell_\text{tar}$.
More ablation studies about normalization methods can be found in Appendix~\ref{app:norm}.

\noindent
\textbf{Filtration with knee point detection.}
When dividing $\Tilde{D}$ into $\Tilde{D}_{e}$ and $\Tilde{D}_{r}$, the threshold $t_\text{f}$ in \eqref{eq:filter} plays a critical role.
For an automatic yet robust choice of $t_\text{f}$, we use the knee point detection mechanism \cite{satopaa2011finding}.
As shown in Figure~\ref{figure:model_inversion_sorted},
there is a knee point in the cumulative distribution of the MMD,
and also the knee point separates $\Tilde{D}_{e}$ and $\Tilde{D}_{r}$ properly
for both {\it \cross{}\!-MNIST} and {\it Truck-CIFAR10}.
Nevertheless, as ours can observe generated samples and whether it is filtered or not, the threshold $t_{\text{f}}$ can be chosen manually if the unlearner wants to.

\begin{table}[t!]
\centering
\begin{tabular}{lccccc}
\toprule
\multicolumn{4}{c}{\multirow{2}{*}{Method}} & \multicolumn{2}{c}{Accuracy (\%)}\\
\cmidrule(lr){5-6}
 & & & & \multicolumn{1}{c}{$D_r (\uparrow)$}  & \multicolumn{1}{c}{$D_e (\uparrow)$}
 \\
\hline
 \multicolumn{4}{c}{Original}                                                & $99.4$ & $  0.0$\\
 \multicolumn{4}{c}{\correct{}}                        & $99.5$ & $100.0$\\
\midrule
                        & $\ell_{tar}$         & $\ell_{aug}$               & $\ell_{bn}$            &         &        \\
\hline
 \multirow{4}{*}{\correcto{}}  &                &                   &               & $98.5 _{\pm 0.1}$ & $ 0.3 _{\pm 0.7}$\\
                        & \checkmark     &                   &               & $93.9 _{\pm 1.0}$ & $98.1 _{\pm 0.7}$\\
                        & \checkmark     & \checkmark        &               & $95.7 _{\pm 0.4}$ & $99.9 _{\pm 0.3}$\\
                        & \checkmark     & \checkmark        & \checkmark    & $99.3 _{\pm 0.0}$ & $100.0 _{\pm 0.0}$\\
\bottomrule
\end{tabular}
\caption{
Ablation study for the prior information ($D_{e}$, augmentation, and batch normalization statistics) on \textit{\cross{}\!-noisyMNIST}.
}
\label{table:ablation-loss}
\end{table}

%% file: sections/conclusion.tex
\section{Conclusion}
\label{sec:conclusion}
 We formulated the comprehensive unlearning problem that covers a wide range of scenarios, including those with few target examples, as well as various purposes: standard unlearning, privacy protection, and mislabel correction.
For this, we proposed the framework, composed of the model inversion, filtration, and relearning steps. Our experiment corroborated the novelty and superiority of each step in the representative yet diverse unlearning scenarios, including the subclass and few-shot unlearning previously infeasible. It also suggested the necessity of more than erasing targets to attain a specific unlearning purpose such as privacy protection or mislabel correction.


\clearpage

%% file: sections/appendix.tex
\section{Implementation Details}
\label{app:detail}
\noindent\textbf{Model architecture.}
For all of our experiments, we use a classifier model architecture, PreAct-ResNet18~\cite{he2016identity}
as the prior works \cite{golatkar2020eternal, chundawat2022zero} use similar ResNet architectures.
As our method works even if it does not have BatchNorm layer as shown in Section~\ref{app:norm}, no specific architecture is required.
Our generator architecture is inspired by DCGAN~\cite{radford2015unsupervised} which has quite simple architecture.
A more complex model could generate better images, but such a simple model still shows promising unlearning performance in experiments.

\noindent\textbf{Detail of $\ell_{\text{tar}}$.}
In the implementation of $\ell_{\text{tar}}$, only batch normalization layers are selected to calculate the mean values of each batch, as done in $\ell_{\text{bn}}$. 
However, unlike $\ell_{\text{bn}}$, it is not restricted to use batch normalization layer only and can be extended to all layers.


\section{Ill-Determined Standard Unlearning}
\label{app:ill}

\begin{figure}[ht!]
    \centering
    \begin{subfigure}{0.48\linewidth}
        \centering
        \includegraphics[width=\linewidth]{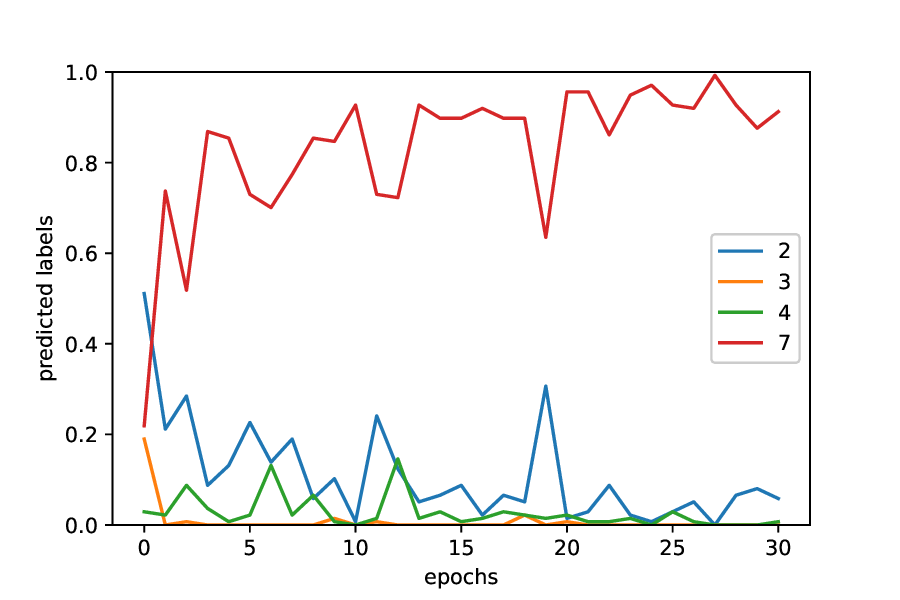}
        \caption{\standard{}}
        \label{figure:app-resnet}
    \end{subfigure}
    \hfill
    \begin{subfigure}{0.48\linewidth}
        \centering
        \includegraphics[width=\linewidth]{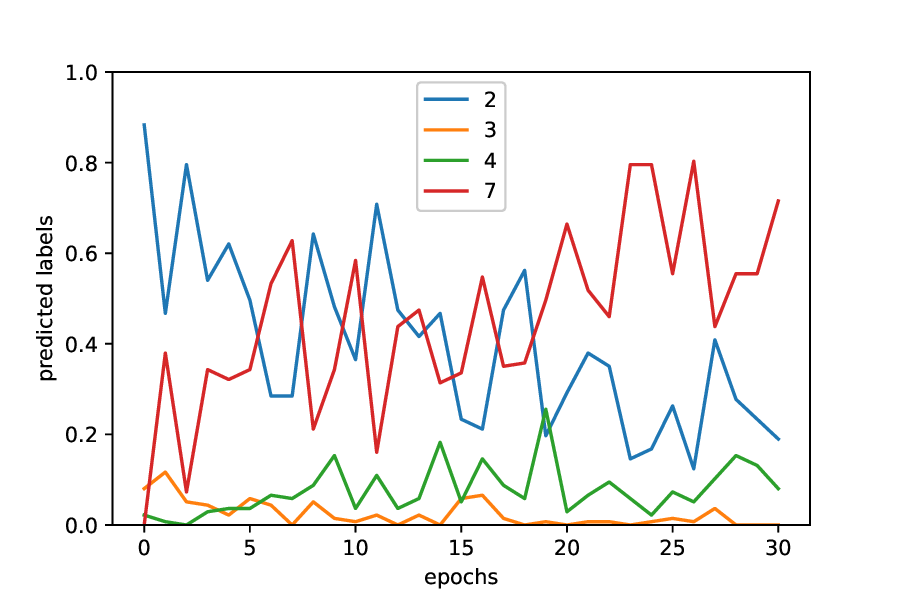}
        \caption{Big learning rate}
        \label{figure:app-biglr}
    \end{subfigure}
    \hfill
    \begin{subfigure}{0.48\linewidth}
        \centering
        \includegraphics[width=\linewidth]{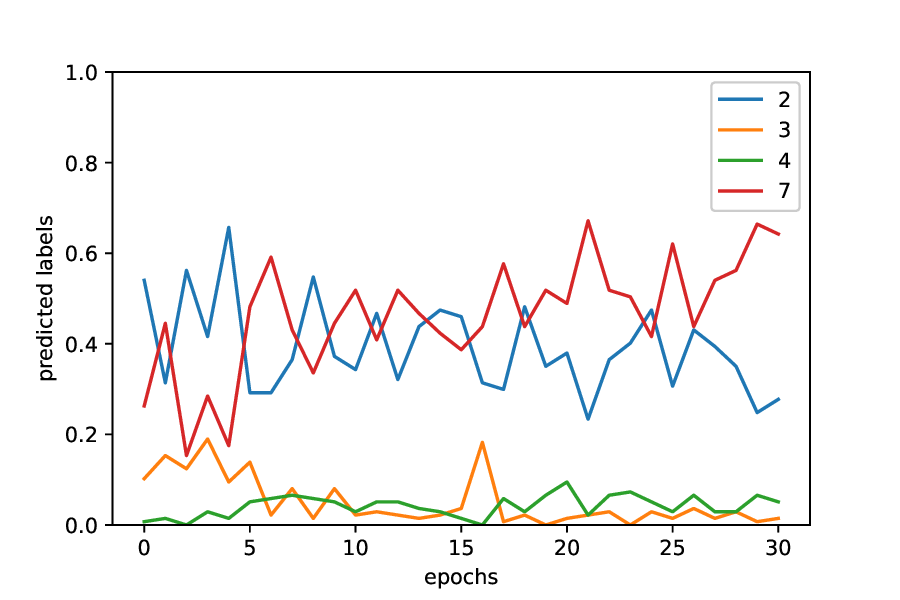}
        \caption{LeNet}
        \label{figure:app-lenet}
    \end{subfigure}
    \hfill
    \begin{subfigure}{0.48\linewidth}
        \centering
        \includegraphics[width=\linewidth]{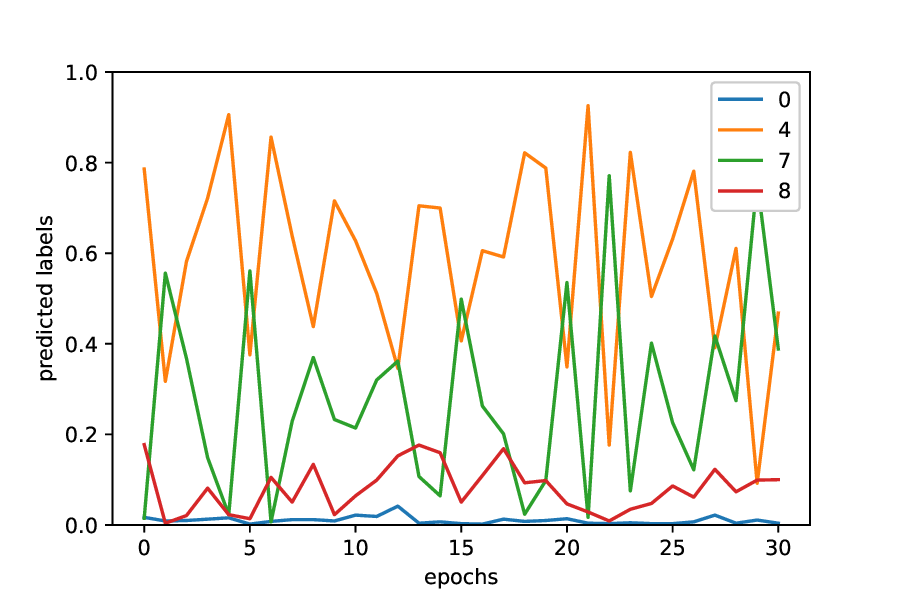}
        \caption{\textit{9-MNIST}}
        \label{figure:app-nine}
    \end{subfigure}
    \caption{
    Predictions for $D_e$ of \standard{}. The predictions often oscillate over training epochs. This shows that the standard unlearning is indeed ill-determined on the model response to the unseen data in $D_e$.
    }
    \label{figure:app-ill}
\end{figure}

\standard{}, trained only with $D_r$ (and without any intention), does not clarify the expected behavior on $D_e$, which can be under-defined and unpredictable.
Thus, we conduct an experiment to show the predictions on $D_e$ in \standard{} on \textit{\cross\!-MNIST} scenario.
In Figure~\ref{figure:app-ill}, we report the predicted labels of $D_e$ over training epochs.
With a big learning rate or simple model architecture in Figure~\ref{figure:app-biglr} and Figure~\ref{figure:app-lenet}, instability of prediction on \cross{} is shown.
Additionally, unlearning 9 shows more vibration, since 9 is similar to both 4 and 7.
Thus, lack of expected behavior on $D_e$ leads to an under-defined oracle and uncertain objective for unlearning, where the intention of the user becomes important.

\section{Unlearning a class}
\label{section:MNIST_class}

We evaluate our algorithm on the class unlearning task, which unlearns a whole class (each of 9, 8, 7) in the MNIST as done in Section~\ref{sec:privacy}.
Our overall results of baseline methods are in Table~\ref{table:987}, all Oracles show similar performance on both $D_r$ and $D_e$ accuracy.
Other baseline methods showed similar performance to Oracle, while {\it Fisher} showed high $D_r$ Accuracy. 
We note that reported previous methods have full training dataset $D$, while our method only uses target data $\Tilde{D}_e$.
Our methods reported slightly less than them but completely erased $D_e$ while maintaining high $D_r$ accuracy.
In addition, our methods achieved performance similar to the oracles on a few-shot setting (3\%) which shows that our method works robustly even in small samples. Ablation study on various ratios of $D_e$ is shown in Appendix~\ref{app:amount}.

\begin{table*}[t!]
\centering
\begin{tabular}{lcccccc}
\toprule
\multirow{2}{*}{Method} & \multicolumn{2}{c}{\textit{9-MNIST} (\%)} & \multicolumn{2}{c}{\textit{8-MNIST} (\%)} & \multicolumn{2}{c}{\textit{7-MNIST} (\%)}\\
\cmidrule(lr){2-7}
 & \multicolumn{1}{c}{$D_r (\uparrow)$}  & \multicolumn{1}{c}{$D_e (\downarrow)$}
 & \multicolumn{1}{c}{$D_r (\uparrow)$}  & \multicolumn{1}{c}{$D_e (\downarrow)$}
 & \multicolumn{1}{c}{$D_r (\uparrow)$}  & \multicolumn{1}{c}{$D_e (\downarrow)$}
 \\
\hline
 Original                           
 & $99.7$ & $99.3$ 
 & $99.7$ & $99.3$ 
 & $99.7$ & $99.3$\\
 \cmidrule(lr){1-7}
 \standard{}  
 & $99.5$ & $0.0$ 
 & $99.5$ & $0.0$ 
 & $99.5$ & $0.0$\\
 \untrained{} 
 & $99.6$ & $0.0$ 
 & $99.3$ & $0.0$ 
 & $99.4$ & $0.0$\\
 \negative{}  
 & $99.5$ & $0.0$ 
 & $99.4$ & $0.0$ 
 & $99.4$ & $0.0$\\
\cmidrule(lr){1-7}
 Incompetent 
 & $99.4_{\pm0.1}$ & $0.0_{\pm0.0}$ 
 & $99.5_{\pm0.2}$ & $0.0_{\pm0.0}$ 
 & $99.4_{\pm0.2}$ & $0.0_{\pm0.0}$ \\
 Neutralized 
 & $99.7_{\pm0.0}$ & $0.0_{\pm0.0}$ 
 & $99.5_{\pm0.1}$ & $0.0_{\pm0.0}$ 
 & $98.9_{\pm0.1}$ & $0.0_{\pm0.0}$ \\
 Fisher 
 & $97.5_{\pm2.4}$ & $96.4_{\pm1.0}$ 
 & $97.9_{\pm1.3}$ & $87.3_{\pm1.4}$ 
 & $98.1_{\pm1.0}$ & $97.0_{\pm1.8}$ \\
 \midrule
 \standardo{}    
 & $98.5_{\pm 0.1}$ & $0.0_{\pm 0.0}$ 
 & $98.4_{\pm 0.2}$ & $0.0_{\pm0.0}$ 
 & $98.3_{\pm 0.3}$ & $0.0_{\pm0.0}$ \\
 \untrainedo{}   
 & $98.5_{\pm 0.4}$ & $0.0_{\pm 0.0}$ 
 & $98.4_{\pm 0.2}$ & $0.7_{\pm1.6}$
 & $97.4_{\pm 0.5}$ & $0.2_{\pm0.2}$ \\
 \negativeo{}    
 & $98.6_{\pm 0.2}$ & $0.0_{\pm 0.0}$ 
 & $98.7_{\pm 0.3}$ & $0.0_{\pm0.0}$ 
 & $98.8_{\pm 0.4}$ & $0.0_{\pm0.0}$ \\
\cmidrule(lr){1-7}
 \standardo{} (3\%)   
 & $98.5_{\pm 0.2}$ & $0.0_{\pm 0.0}$ 
 & $97.9_{\pm 0.6}$ & $0.0_{\pm0.0}$ 
 & $98.3_{\pm 0.4}$ & $1.2_{\pm2.8}$ \\
 \untrainedo{} (3\%)   
 & $98.1_{\pm 0.5}$ & $0.0_{\pm 0.0}$ 
 & $98.1_{\pm 0.2}$ & $0.2_{\pm0.2}$ 
 & $97.3_{\pm 1.1}$ & $1.0_{\pm1.1}$ \\
 \negativeo{} (3\%)   
 & $98.8_{\pm 0.1}$ & $0.0_{\pm 0.0}$ 
 & $98.3_{\pm 0.3}$ & $0.0_{\pm 0.0}$ 
 & $99.1_{\pm 0.2}$ & $0.3_{\pm 0.3}$ \\

\bottomrule
\end{tabular}
\caption{
 Unlearned models for unlearning each class of 9, 8, and 7 in MNIST, denoted by {\it 9-MNIST}, {\it 8-MNIST} and {\it 7-MNIST}, respectively. 
 We report the mean with standard deviation over 5 random instances.
}
\label{table:987}
\end{table*}

\section{Ablation Studies}
\subsection{Ablation study on the amount of given $D_{e}$}
\label{app:amount}

Here, we study the ratio of given $D_{e}$; 100\%, 30\%, 10\%, 3\% and single image.
Oracle reported results based on the ratio of $D_e$, which means the remaining ratio is included in $D_r$.
The experiments are conducted in the setting from Section~\ref{sec:privacy}, which unlearns \cross{} in MNIST.
Table~\ref{table:app-amount} shows that all three of our methods relatively robustly keep the accuracy of $D_r$.
The cases of \standardo{} and  \untrainedo{} do not show a tendency according to the amount of De because oracle is not stable, as shown in Section~\ref{app:ill}.
\negativeo{} shows high accuracy on $D_e$ when few samples are used, as the filter failed to filtrate many \cross{}, which leads to learn \cross{} as 7.
However, \standard{}, \untrained{}, and \negative{} show similar accuracy with Original when only portion of $D_e$ is given, which assumes remaining target data is included in $D_r$.

\begin{table}[t!]
\centering
\begin{tabular}{lcc}
\toprule
\multirow{2}{*}{Method} & \multicolumn{2}{c}{Accuracy (\%)}\\
\cmidrule(lr){2-3}
 & \multicolumn{1}{c}{$D_r$}  & \multicolumn{1}{c}{$D_e$}
 \\
\hline
 $\text{Original}$                                & $99.69$ & $98.54$ \\
 \midrule
 \standard{} (100\%)          & $99.50$ & $74.45$ \\
 \standard{} (30\%)           & $98.78$ & $94.16$ \\
 \standard{} (10\%)           & $99.16$ & $99.27$ \\
 \standard{} (3\%)            & $99.04$ & $100.00$ \\
 \standard{} (1)              & $99.33$ & $97.81$ \\

  \cmidrule(lr){1-3}
 
 \untrained{} (100\%)         & $99.58$ & $18.98$ \\
 \untrained{} (30\%)          & $99.18$ & $97.81$ \\
 \untrained{} (10\%)          & $99.33$ & $95.62$ \\
 \untrained{} (3\%)           & $99.41$ & $99.27$ \\
 \untrained{} (1)             & $99.36$ & $97.81$ \\

 \cmidrule(lr){1-3}
 
 \negative{} (100\%)          & $98.05$ & $ 0.00$ \\
 \negative{} (30\%)           & $99.26$ & $95.62$ \\
 \negative{} (10\%)           & $99.28$ & $98.54$ \\
 \negative{} (3\%)            & $99.48$ & $99.27$ \\
 \negative{} (1)              & $99.31$ & $99.27$ \\

 \midrule
 \standardo{} (100\%)            & $99.26 _{\pm 0.30}$ & $81.46 _{\pm 20.1}$ \\
 \standardo{} (30\%)             & $99.26 _{\pm 0.22}$ & $83.21 _{\pm 18.9}$ \\
 \standardo{} (10\%)             & $99.31 _{\pm 0.45}$ & $81.02 _{\pm 22.7}$ \\
 \standardo{} (3\%)              & $99.56 _{\pm 0.05}$ & $90.80 _{\pm 2.61}$ \\
 \standardo{} (1)                & $99.53 _{\pm 0.00}$ & $76.78 _{\pm 1.58}$ \\
 \cmidrule(lr){1-3}
 \untrainedo{} (100\%)           & $99.44 _{\pm 0.08}$ & $77.66 _{\pm 2.45}$ \\
 \untrainedo{} (30\%)            & $99.20 _{\pm 0.09}$ & $32.70 _{\pm 4.60}$ \\
 \untrainedo{} (10\%)            & $99.17 _{\pm 0.13}$ & $59.12 _{\pm 12.7}$ \\
 \untrainedo{} (3\%)             & $99.23 _{\pm 0.10}$ & $73.72 _{\pm 8.68}$ \\
 \untrainedo{} (1)               & $99.21 _{\pm 0.11}$ & $39.12 _{\pm 5.94}$ \\
 \cmidrule(lr){1-3}
 \negativeo{} (100\%)            & $98.45 _{\pm 0.46}$ & $ 2.04 _{\pm 0.61}$ \\
 \negativeo{} (30\%)             & $98.39 _{\pm 1.55}$ & $ 3.80 _{\pm 1.20}$ \\
 \negativeo{} (10\%)             & $98.48 _{\pm 1.30}$ & $ 4.96 _{\pm 2.21}$ \\
 \negativeo{} (3\%)              & $98.08 _{\pm 0.13}$ & $11.68 _{\pm 0.52}$ \\
 \negativeo{} (1)                & $98.76 _{\pm 1.28}$ & $20.44 _{\pm 0.89}$ \\
\bottomrule
\end{tabular}
\vspace{1em}
\caption{
Unlearning performance at different $D_e$ amounts.
We report the result of {\it \cross{}\!-MNIST} in Section~\ref{sec:privacy} to erase the subclass \cross in MNIST.
}
\label{table:app-amount}
\end{table}

\subsection{Ablation study on normalization methods}
\label{app:norm}

Table~\ref{table:ablation-loss} shows that $\ell_\text{bn}$ is the strong performance-enhancing loss in our method.
However, recently, models that use various normalization methods such as LayerNorm and InstanceNorm are also been widely used.
Thus, we conduct the experiment on the other normalized method without $\ell_{bn}$, which follows the same setting with Section~\ref{section:MNIST_class} for the \negative{} and \negativeo{}.
To compare the normalization method only, we just change the normalization layer of the model architecture from BatchNorm to LayerNorm or InstanceNorm.
As shown in Table~\ref{table:app-norm}, our methods had a slight drop in performance on $D_r$ but can erase almost all $D_e$ with even other normalization methods. 

\begin{table}[t!]
\centering
\begin{tabular}{lcc}
\toprule
\multirow{2}{*}{Method} & \multicolumn{2}{c}{Accuracy (\%)}\\
\cmidrule(lr){2-3}
 & \multicolumn{1}{c}{$D_r$}  & \multicolumn{1}{c}{$D_e$}
 \\
\hline
 $\text{Original}_{\text{BN}}$            & $99.71$ & $99.31$ \\
 $\text{Oracle}_{\text{BN}}$              & $99.52$ & $0.00$ \\
 $\text{Ours}_{\text{BN}}$                & $98.61 _{\pm 0.23}$ & $0.00 _{\pm 0.00}$ \\
 $\text{Ours}_{\text{BN}} (3\%)$          & $98.80 _{\pm 0.14}$ & $0.00 _{\pm 0.00}$ \\
 \cmidrule(lr){1-3}
 $\text{Original}_{\text{LN}}$            & $99.31$ & $98.81$ \\
 $\text{Oracle}_{\text{LN}}$              & $99.52$ & $ 0.00$ \\ 
 $\text{Ours}_{\text{LN}}$                & $97.95 _{\pm 0.24}$ & $0.00 _{\pm 0.00}$ \\ 
 $\text{Ours}_{\text{LN}} (3\%)$          & $97.76 _{\pm 0.51}$ & $0.00 _{\pm 0.00}$ \\
  \cmidrule(lr){1-3}
 $\text{Original}_{\text{IN}}$            & $99.11$ & $99.69$ \\
 $\text{Oracle}_{\text{IN}}$              & $99.59$ & $ 0.00$ \\
 $\text{Ours}_{\text{IN}}$                & $97.51 _{\pm 0.78}$ & $0.06 _{\pm 0.05}$ \\
 $\text{Ours}_{\text{IN}} (3\%)$          & $96.35 _{\pm 0.47}$ & $1.19 _{\pm 0.56}$ \\
\bottomrule
\end{tabular}
\vspace{1em}
\caption{
Unlearning performance for different normalization methods on the task which unlearning 9 in MNIST.
All of the experiments are conducted in a negative setting.
$\text{*}_\text{BN}$, $\text{*}_\text{LN}$, and $\text{*}_\text{IN}$ are the methods when the baseline classifier uses BatchNorm, LayerNorm, and Instance Norm.
}
\label{table:app-norm}
\end{table}

\subsection{Ablation study on scrubbing} 
\label{app:scrub}

For standard unlearning by retraining, \citet{Kim_2022_CVPR} show that scrubbing before retraining decreases unlearning epochs.
Thus, we study scrubbing for \standardo{}.
The experiments setting follows Section~\ref{section:MNIST_class}, \textit{9-MNIST}.
Table~\ref{table:app-scrub} shows that $\text{Ours}_\text{std$-$scrub}$ still remember $D_e$ even after unlearning,
which emphasizes that scrubbing is essential for \standardo{} to erase $D_e$.

\begin{table}[t!]
\centering
\begin{tabular}{lcc}
\toprule
\multirow{2}{*}{Method} & \multicolumn{2}{c}{Accuracy (\%)}\\
\cmidrule(lr){2-3}
 & \multicolumn{1}{c}{$D_r$}  & \multicolumn{1}{c}{$D_e$}
 \\
\hline
 Original                                 & $99.71$ & $99.31$ \\
 \standard{}        & $99.52$ & $ 0.00$ \\
 \cmidrule(lr){1-3}
 \standardo{}                            & $98.53 _{\pm 0.11}$ & $ 0.00 _{\pm 0.00}$ \\
 $\text{Ours}_{\text{std$-$scrub}}$            & $99.18 _{\pm 0.21}$ & $96.14 _{\pm 1.37}$ \\
 
 \cmidrule(lr){1-3}
 \standardo (3\%)                      & $98.53 _{\pm 0.15}$ & $ 0.00 _{\pm 0.00}$ \\
 $\text{Ours}_{\text{std$-$scrub}} (3\%)$      & $99.35 _{\pm 0.06}$ & $96.71 _{\pm 0.41}$ \\

\bottomrule
\end{tabular}
\vspace{1em}
\caption{
Unlearning performance without scrubbing for \standardo{} on \textit{9-MNIST} scenario.
$\text{Ours}_\text{std$-$scrub}$ denotes \standardo{} without scrubbing.
}
\label{table:app-scrub}
\end{table}

\section{Privacy on Class unlearning}
\label{app:privacy}
In Table~\ref{table:app_privacy}, we report the result of ASR and L2 norm on \textit{9-MNIST} scenario.
Similar to Table~\ref{table:privacy}, \untrained{}, Incompetent, \standardo{}, and \untrainedo{} show relatively small ASR than others.
Moreover, overall ASR in Table~\ref{table:app_privacy} is smaller than that in Table~\ref{table:privacy}, as it requires erasing a large amount of information.
For the same reason, the difference in L2 norms between $D_r$ and $D_e$ also widened.

\begin{table}[t!]
\centering
\begin{tabular}{lcccc}
\toprule
\multirow{2}{*}{Method} & \multicolumn{2}{c}{ASR (\%)} & \multicolumn{2}{c}{L2 norm}\\
\cmidrule(lr){2-5}
 & \multicolumn{1}{c}{$D_r$}  & \multicolumn{1}{c}{$D_e$}
 & \multicolumn{1}{c}{$D_r$}  & \multicolumn{1}{c}{$D_e$}
 \\
\hline
 Original                                 
 & $100.0$ & $100.0$ 
 & $20.84 _{\pm 3.34}$ & $18.40 _{\pm 2.88}$
 \\
 \cmidrule(lr){1-5}
 \standard{}       
 & $100.0$ & $100.0$ 
 & $22.10 _{\pm 4.33}$ & $15.45 _{\pm 1.76}$
 \\
 \untrained{}       
 & $98.4$ & $3.2$ 
 & $25.04 _{\pm 6.94}$ & $ 5.05 _{\pm 0.59}$
 \\
 \negative{}       
 & $100.0$ & $100.0$ 
 & $24.19 _{\pm 4.20}$ & $15.92 _{\pm 2.00}$
 \\
 \cmidrule(lr){1-5}
 Incompetent 
 & $98.3$ & $12.8$ 
 & $17.57 _{\pm 5.25}$ & $ 2.27 _{\pm 1.17}$
 \\
 Neutralized 
 & $99.9$ & $55.3$ 
 & $22.59 _{\pm 4.45}$ & $13.40 _{\pm 2.12}$
 \\
 Fisher 
 & $100.0$ & $100.0$ 
 & $19.54 _{\pm 3.20}$ & $17.72 _{\pm 2.44}$
 \\
 \midrule
 \standardo{}         
 & $99.7$ & $19.8$ 
 & $23.44 _{\pm 4.89}$ & $ 7.34 _{\pm 2.04}$
 \\
 \untrainedo{}         
 & $99.9$ & $13.2$ 
 & $23.79 _{\pm 5.87}$ & $ 4.80 _{\pm 2.57}$
 \\
 \negativeo{}         
 & $99.8$ & $60.2$ 
 & $20.44 _{\pm 3.61}$ & $11.77 _{\pm 2.00}$
 \\
 \cmidrule(lr){1-5}
 \standardo{} (3\%)  
 & $99.7$ & $19.9$ 
 & $21.21 _{\pm 4.29}$ & $ 7.71 _{\pm 2.53}$
 \\
 \untrainedo{} (3\%)  
 & $99.9$ & $16.8$ 
 & $21.88 _{\pm 5.85}$ & $ 4.38 _{\pm 2.33}$
 \\
 \negativeo{} (3\%)  
 & $99.9$ & $55.0$ 
 & $19.57 _{\pm 4.94}$ & $10.74 _{\pm 2.40}$
 \\

\bottomrule
\end{tabular}
\vspace{1em}
\caption{
Attack Success Rate and L2 norm results to check privacy on De for unlearning class 9 in MNIST. 
Attack Success Rate reports the result of the membership inference attack method. 
L2 norm is calculated on the latent space.
}
\label{table:app_privacy}
\end{table}

\section{Formal Definition of $\ell_\text{tv}$ and $\ell_\text{div}$}
\label{app:def}

Here, we give a formal definition of losses used for model inversion, $\ell_{\text{tv}}$ and $\ell_{\text{div}}$ from \ref{sec:method-inversion}.

\smallskip
\noindent
\textbf{Diversity loss.} 
GANs often suffer from mode-collapsing \cite{goodfellow2014generative}. 
In our case, having a few collapsed samples per label can potentially make the unlearning unstable since the coverage of the generated samples is limited.
To overcome the issue of mode-collapse, we propose a modification of diversity loss proposed in \cite{yoo2019knowledge}.
Let $fe:\mathcal{X}\rightarrow \mathbb{R}^D$ be a feature extractor, which is the target classifier except for the last fully connected layer. Our diversity loss is defined as
\begin{equation}
\begin{multlined}
    \ell_{\text{div}} (B) = 
\exp\left(  
    -
    \sum_{ 
        \substack{
        (z_i, y) \in B \\
        (z_j, y) \in B
        }
    }
    \lVert{z_{i} - z_{j}}\rVert_{2}
    \cdot
    d_{fe}
    \right)
    ,
\end{multlined}
\end{equation}

where $z_i$, $z_j$ are two randomly sampled noises, 
and $d_{fe}$ denotes the feature-wise distance between two generated samples, 
where the distance is measured with $l_{1}$ norm. 
The feature is an output of convolution layers of the target classifier.
In \cite{yoo2019knowledge}, the pixel-wise distance between $G(z, y)$ is used instead of the feature-wise distance.

\smallskip
\noindent
\textbf{Total variation loss.} 
One of the common prior on 
natural images is that they have local smoothness, i.e., a small 
total variation.
To enforce the local smoothness of a generated image, we have used the total variation loss defined as:
\begin{align}
    \ell_{\text{tv}}(B) = 
    \sum_{(z, y)\in B}{
        \sum_{(i, j)}{
            \sum_{(i', j') \in \delta(i, j)}
            {
                \lVert{\Tilde{x}_{(i, j)} - \Tilde{x}_{(i', j')}}\rVert_{2}^{2}
            }
        }
    }\;,
\end{align}
where $\delta(i, j)$ indicates a set of pixels adjacent to pixel $(i, j)$.
The subscript indexes the generated pixel at a corresponding location.